\documentclass[letterpaper, 10 pt, conference]{ieeeconf}  

\IEEEoverridecommandlockouts                              

\makeatletter
\let\NAT@parse\undefined
\newcommand{\ie}{\emph{i.e.}\@ifnextchar.{\!\@gobble}{}}
\newcommand{\eg}{\emph{e.g.}\@ifnextchar.{\!\@gobble}{}}
\newcommand{\etc}{etc\@ifnextchar.{}{.\@}}
\makeatother
\usepackage[numbers,sort&compress]{natbib}

\usepackage[pdftex]{graphicx}
\usepackage{epsfig} 
\usepackage{times} 
\usepackage{amsmath} 
\usepackage{amssymb}  
\usepackage{lipsum}
\usepackage{pifont}
\usepackage{algpseudocode}
\usepackage{mathtools}
\usepackage{color}
\usepackage{subfigure}
\usepackage[table,xcdraw]{xcolor}
\usepackage{adjustbox}
\usepackage{multicol, multirow}
\usepackage{booktabs}
\usepackage[normalem]{ulem}
\usepackage{balance}
\usepackage{hyperref}
 \hypersetup{
     colorlinks=true,
     linkcolor=blue,
     filecolor=magenta,
     urlcolor=blue,
     citecolor=black
 }

\usepackage{rpm_SIunits}
\usepackage{rpm_acronyms}
\usepackage{rpm_math}
\usepackage{rpm_misc}

\usepackage{soul}

\title{\LARGE \bf
Ephemerality meets LiDAR-based Lifelong Mapping
}

\author{Hyeonjae Gil$^{1\dagger}$, Dongjae Lee$^{1\dagger}$, Giseop Kim$^{2}$, and Ayoung Kim$^{1*}$
\thanks{$^\dagger$Equal contribution, $^*$Corresponding author.}%
\thanks{$^\ddagger$This work was supported by the National Research Foundation of Korea (NRF) grant funded by the Korea government (MSIT)(No. RS-2024-00461409), and in part by the Robotics and AI (RAI) Institute.}%
\thanks{$^{1}$H. Gil, D. Lee, and A. Kim are with the Department of Mechanical Engineering, Seoul National University, Seoul, S. Korea
        {\tt\small [h.gil, pur22, ayoungk]@snu.ac.kr}}%
\thanks{$^{2}$G.Kim is with Vision Group of NAVER LABS, Seongnam, Gyeonggi-do, 13561, S. Korea
        {\tt\small giseop.kim@naverlabs.com}}%
}

\begin{document}

\maketitle
\thispagestyle{empty}
\pagestyle{empty}

\begin{abstract}

Lifelong mapping is crucial for the long-term deployment of robots in dynamic environments.
In this paper, we present ELite, an ephemerality-aided LiDAR-based lifelong mapping framework which can seamlessly align multiple session data, remove dynamic objects, and update maps in an end-to-end fashion.
Map elements are typically classified as static or dynamic, but cases like parked cars indicate the need for more detailed categories than binary.
Central to our approach is the probabilistic modeling of the world into two-stage \textit{ephemerality}, which represent the transiency of points in the map within two different time scales.
By leveraging the spatiotemporal context encoded in ephemeralities, ELite can accurately infer transient map elements, maintain a reliable up-to-date static map, and improve robustness in aligning the new data in a more fine-grained manner. 
Extensive real-world experiments on long-term datasets demonstrate the robustness and effectiveness of our system.
The source code is publicly available for the robotics community: \href{https://github.com/dongjae0107/ELite}{https://github.com/dongjae0107/ELite}.

\end{abstract}
\section{INTRODUCTION}
\label{sec:intro}

Over the past decade, \ac{LiDAR}-based mapping has significantly advanced \cite{zhang2014loam, shan2018lego, shan2020lio, xu2022fast}, increasing the demand for long-term deployment of such systems in various fields, including urban areas or construction sites \cite{lee2024lidar}. 
These environments are inherently dynamic; objects frequently move, and layouts change. 
To handle these dynamics, continuously revisiting and maintaining the map of the environment---lifelong mapping---is required.

LiDAR-based lifelong mapping has gained interest relatively recently compared to the visual domain \cite{09towards, cramariuc2022maplab, elvira2019orbslam}.
Long-term mapping pipelines for static map \cite{pomerleau2014long} or semantic map \cite{sun2018recurrent} construction were suggested, but the standard framework for lifelong mapping was absent.
Recently, the modular lifelong mapping frameworks \cite{kim2022lt, yang2024lifelong} were suggested. 
They process the session---a set of point clouds and poses---as an input and focus on the \textit{inter-session changes} for efficient map management and incremental update.

\begin{figure}[!t]
    \centering
    \includegraphics[width=0.95\linewidth]{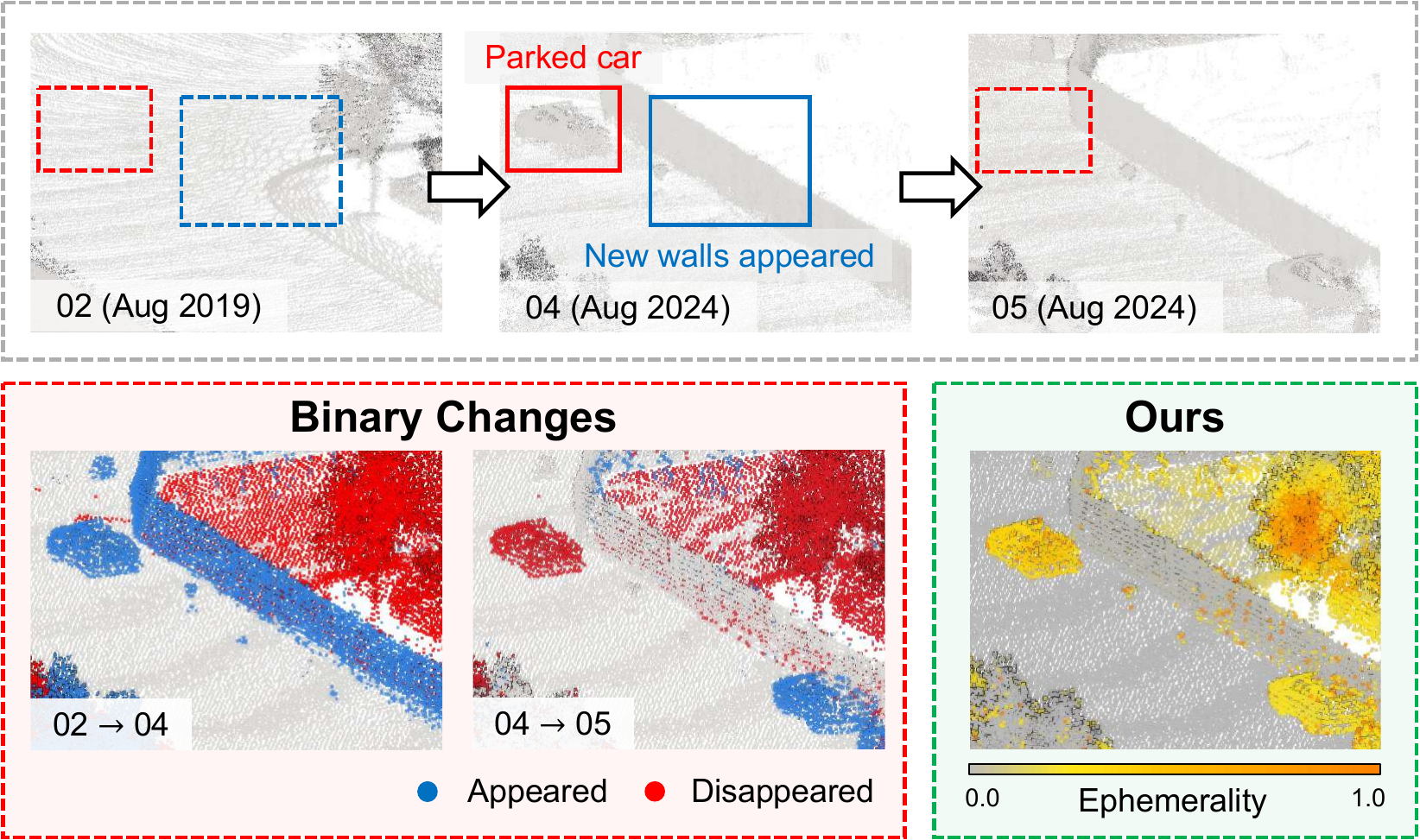}
    \caption{
    An example scene from three \texttt{KAIST} sequences \cite{kim2020mulran, jung2023helipr} with newly appeared walls and parked cars. 
    Representing changes in a simple binary manner, existing methods treat both the car and the wall as static objects. 
    Our proposed system leverages two-stage ephemerality to differentiate parked cars as ephemeral objects and walls as persistent changes based on their ephemerality scores.
    }
    \label{fig:figure_1}
    \vspace{-5mm}
\end{figure}

These changes have been modeled as binary (\ie, appearing or disappearing), leading to a binary classification of map elements (\ie, static or dynamic). 
The inherent limitation of this approach is its inability to differentiate between long-term gradual changes and short-term ephemeral variations. 
An example is illustrated in \figref{fig:figure_1}, where two objects appear on the new map: one represents a persistent change (new walls), and the other reflects a relatively short-term variation (parked cars).
Unfortunately, both are categorized into a single static class in existing methods \cite{kim2022lt, yang2024lifelong}; yet, our approach can distinguish them based on their ephemerality score. 

The key idea is that changes in real-world are gradual. Detailing changes beyond binary categorization, we introduce \textit{ephemerality} as the core concept of our lifelong mapping framework.
Ephemerality represents the likelihood of a point being transient or persistent. 
Previous literature \cite{09towards, mcmanus2013distraction, barnes2018driven} has commonly treated ephemeral objects the same as dynamic ones (\eg, pedestrians, moving vehicles) in short-term contexts. 
In this paper, we extend the focus to long-term perspectives, elaborating dynamic objects with details from ephemeral variation to gradual map evolution.

This paper builds on the modular framework of LT-mapper \cite{kim2022lt}.
Unlike previous approaches \cite{kim2022lt, yang2024lifelong} with three independent modules, ours facilitates seamless integration of each module with ephemerality, which permeates the entire pipeline and enhances both per-module and overall performance.
In doing so, we infer a two-stage ephemerality with different time scales to express the subtle differences.
This allows us to represent changes between sessions in a more fine-grained manner than traditional binary approaches. 
In addition, leveraging spatiotemporal context, we can accurately distinguish meaningful changes from those resulting from errors and use them for effective map updates. 
The contributions of our system are as follows.
\begin{itemize}
    \item We introduce a two-stage ephemerality concept---local and global---to capture short-term and long-term changes, respectively.
    This approach extends beyond binary static/dynamic classification by distinguishing truly persistent changes from transient variations.

    \item
    We propose ELite, a LiDAR-based lifelong mapping framework that incorporates ephemerality into each module. 
    Our approach uses ephemerality to guide map alignment, prioritize meaningful map updates, and robustly detect evolving structures over time.

    \item 
    ELite maintains three types of maps: \textit{a lifelong map} capturing spatiotemporal history, an adjustable \textit{static map} filtering out ephemeral clutter, and an object-oriented \textit{delta map} highlighting changed components. These representations enable flexible usage based on different requirements and time horizons.

    \item Each module within ELite has been thoroughly evaluated, showing superior performance compared to the baselines. All codes and related softwares are open-sourced for the community.
\end{itemize}
\section{RELATED WORK}
\label{sec:realtedwork}

\subsection{Change Detection}
In the 3D change detection literature, many methods adopt an object-centric approach. For instance, \citeauthor{schmid2022panoptic} \cite{schmid2022panoptic} and \citeauthor{langer2020robust} \cite{langer2020robust} define and manage changes based on panoptic or semantic segmentation. However, these strategies often rely on neural networks trained on large amounts of labeled data, which may be infeasible for diverse or unstructured outdoor environments.
Alternatively, several approaches \cite{rowell2024lista, adam2022objects} leverage geometric changes as a prior for reconstructing object-level differences between sessions. Although effective, they assume that changes occur in discrete, object-wise units---an assumption that may break down in highly dynamic outdoor settings, such as construction sites where sand or soil is incrementally added. To address this limitation, we detect changes at the point cloud level and maintain point-wise ephemerality, thereby accommodating continuous or non-discrete changes. This allows us to handle a broader range of real-world scenarios and move beyond a binary changed/not-changed classification paradigm.

\subsection{LiDAR-based Lifelong Mapping}
LiDAR-based lifelong mapping has dealt with scalability \cite{lazaro2018efficient, zhao2021general, kurz2021geometry} or predictability \cite{krajnik2017fremen}, but most of them were demonstrated in two-dimensional spaces.
\citeauthor{pomerleau2014long} \cite{pomerleau2014long} suggested 3D map maintenance pipeline, but they assumed accurate registration and lacked the ability to revert the updates.
Recently, LT-mapper \cite{kim2022lt} suggested the modular approach for lifelong mapping with the following three modules.

\subsubsection{Multi-session map alignment} 
Aligning a point cloud map is often viewed as a registration problem \cite{yin2023automerge, 24frame, yang2024lifelong}. 
However, relying solely on simple rigid-body transformations can introduce alignment errors when the mapped region expands \cite{rowell2024lista}.
To address these challenges, multi-session \ac{PGO} frameworks \cite{kim2022lt, rowell2024lista} have been proposed, but they still face local inconsistencies in large-scale environments. 

\subsubsection{Dynamic object removal} 
Geometry-based methods discretize the environment using voxels \cite{hornung2013octomap, schmid2023dynablox, duberg2024dufomap}, range images \cite{kim2020remove}, bins \cite{lim2021erasor, lim2023erasor2}, or matrices \cite{jia2024beautymap}.
However, these methods are constrained by grid resolution, risking inaccuracies when a single cell contains both static and dynamic points. 
Learning-based methods \cite{pfreundschuh2021dynamic, mersch2022receding, sun2022efficient} can also be effective but typically require extensive labeled datasets to maintain robust performance in unfamiliar scenarios.

\subsubsection{Map update} 
LT-mapper \cite{kim2022lt} and \citeauthor{yang2024lifelong} \cite{yang2024lifelong} detect changes between sessions and update the existing map accordingly. 
They save the changed points and use a version control system \cite{spinellis2012git} that allows manual rollbacks \cite{holoch2022detecting} to previous map via simple arithmetic operations.
Unfortunately, these methods treat changes as binary, which dilutes meaningful changes with outliers from various error sources.

Extending the modular nature, ELite addresses the drawbacks in each of the three modules by introducing ephemerality as a unifying concept throughout the pipeline. It identifies static and persistent regions during multi-session alignment, removes dynamic objects without discretization, and prioritizes meaningful changes for map updates by leveraging contextual information. This integrated use of ephemerality helps ensure more accurate and robust lifelong mapping in real-world, continuously evolving environments.
\section{METHOD}
\label{sec:method}

\subsection{System Overview}
\label{subsec:system_overview}

\begin{figure}[!t]
    \centering
    \includegraphics[width=0.95\linewidth]{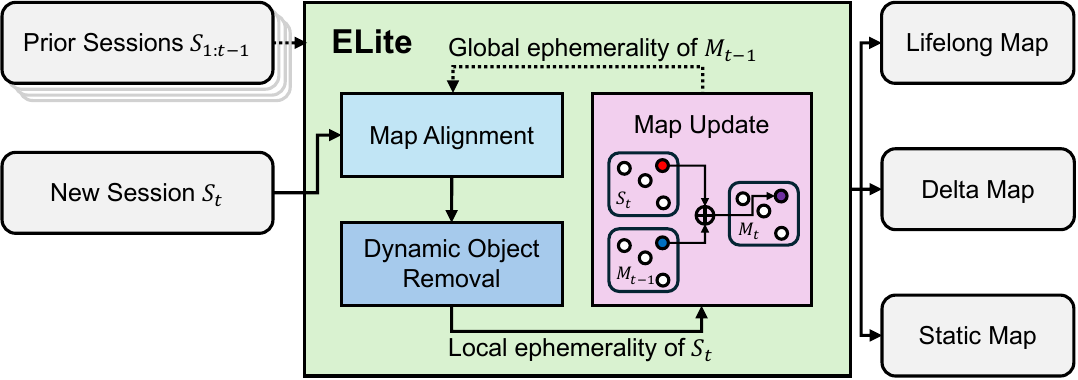}
    \caption{Overview of the ELite system pipeline. Given multiple input sessions, ELite updates the map by estimating local ephemerality within each session and updating global ephemerality across sessions. The system operates through three modules: multi-session map alignment, dynamic object removal, and map update. Additionally, the system manages three types of maps: a lifelong map, a delta map, and a static map.}
    \label{fig:system_overview}
    \vspace{-5mm}
\end{figure}

ELite manages two stages of ephemerality: \emph{local ephemerality} ($\epsilon_{l}$) and \emph{global ephemerality} ($\epsilon_{g}$).
Here, $\epsilon_{l}$ reflects the probability of a point being dynamic within a single session (\eg, moving cars have higher $\epsilon_{l}$ than parked cars), while $\epsilon_{g}$ captures the long-term likelihood of a point being transient (\eg, a car repeatedly parked in the same location exhibits a higher $\epsilon_{g}$ than a permanent building).

\figref{fig:system_overview} provides an overview of the system.
Starting from the base map $\mathcal{M}_1$, which is built directly from the first session $\mathcal{S}_1$ (see \secref{subsec:dynamic_removal}), our lifelong mapper $L(\cdot)$ incrementally updates the previous lifelong map $\mathcal{M}_{t-1}$ using the new session $\mathcal{S}_t$:
\begin{equation}
\mathcal{M}_{t}= \left\{
\begin{aligned}
&{\hat{\mathcal{M}}_{S_t}}, \quad &t=1 \\
&L(\mathcal{M}_{t-1}, S_{t}), \quad &t>1 \\
\end{aligned}
\right.
\end{equation}

The \emph{lifelong map} is a point cloud in which each point contains $(x,y,z, \epsilon_{g})$.  
A session $\mathcal{S} = \left\{ \left( \mathcal{P}_i, \mathbf{T}_i \right) \right\}_{i=1}^N$ is a set of scans and their corresponding poses in a local coordinate frame, obtained from any positioning system (\eg. LiDAR odometry \cite{xu2022fast} or SLAM \cite{shan2020lio}).

First, ELite aligns the new session to the existing lifelong map by finding the scan-wise optimal transformations (\secref{subsec:map_alignment}).  
Next, it estimates local ephemerality $\epsilon_{l}$ for each point via point-wise ephemerality propagation in the \emph{dynamic object removal} module and then discards dynamic points (\secref{subsec:dynamic_removal}).  
Finally, leveraging the estimated local ephemerality, the \emph{map update} module classifies map points into multiple categories and applies category-specific update rules to compute the new global ephemerality $\epsilon_{g}$ (\secref{subsec:map_update}).  
Over repeated updates, $\epsilon_{g}$ acts as a reliable weight for aligning newly acquired sessions, thereby enhancing system robustness during extended operation.

\subsection{Multi-session Map Alignment}
\label{subsec:map_alignment}

\begin{figure}[!t]
    \centering
    \includegraphics[width=0.95\linewidth]{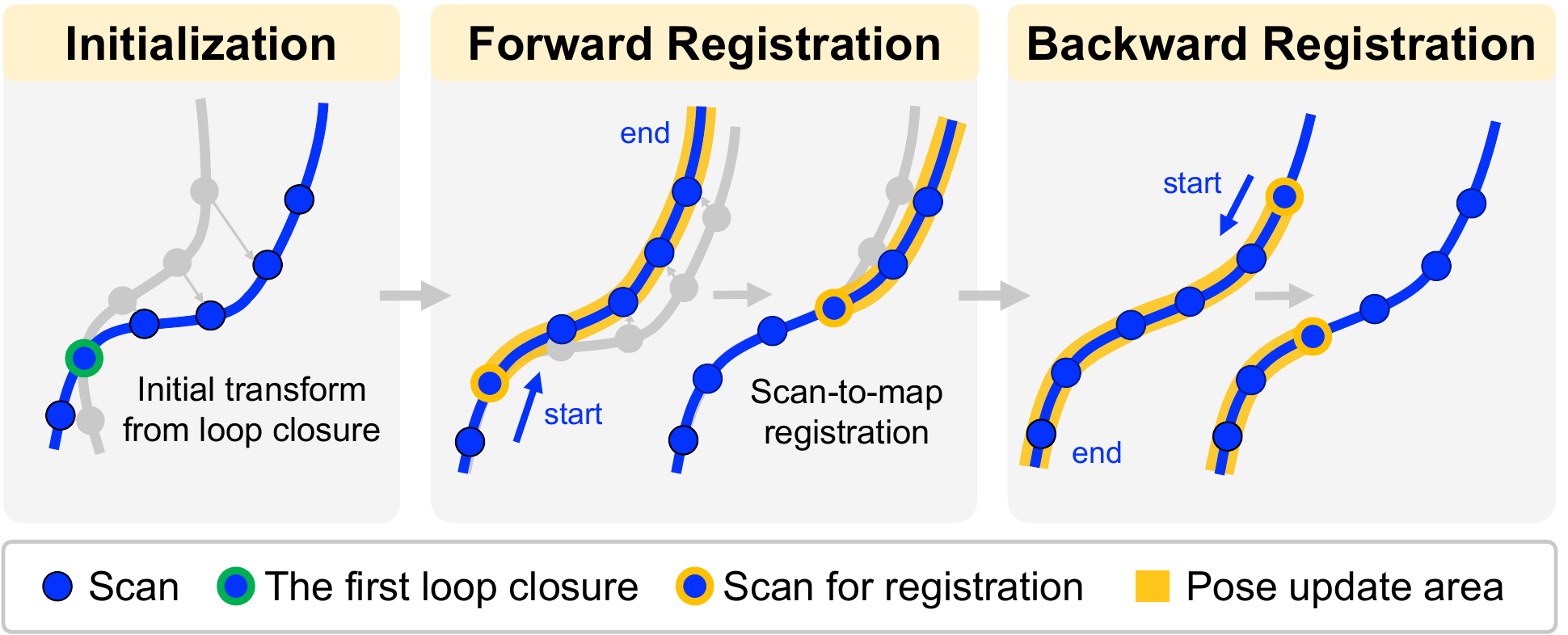}
    \caption{
    Illustration of our map alignment module, which begins by aligning poses using the initial transform from the first loop closure candidate. It then refines alignment in two stages: forward and backward. By iterating through scans in both directions, the module updates poses via scan-to-map ICP registration, ensuring global and local consistency in the final pose estimates.
    }
    \label{fig:zipper}
    \vspace{-3mm}
\end{figure}

The goal of an alignment module is to find the optimal transformation \( \mathbf{T}'_i \) for each scan $\mathcal{P}_i$ in a session and pass the aligned session \(\mathcal{S'} = \left\{ \left( \mathcal{P}_i, \mathbf{T}'_i \right) \right\}_{i=1}^N\) to subsequent modules.

\figref{fig:zipper} illustrates our alignment pipeline. 
We first select a loop pair $(\mathcal{P}^{t-1}_{m}, \mathcal{P}^{t}_{s})$ from two sessions ($\mathcal{S}_{t-1}$ and $\mathcal{S}_{t}$) via Scan Context \cite{kim2021scan} to estimate an initial transformation \( \mathbf{T}^{\text{init}} \).  
We then refine \( \mathbf{T}^{\text{init}} \) in a \emph{forward} pass over indices \(i \in [s, N]\) using \ac{GICP} \cite{koide2021voxelized}:
\begin{equation}
\vspace{-1mm}
\mathbf{T}^{\text{fwd}}_i = 
\prod_{j=s}^{i} \mathbf{T}_j^{\text{ICP}} \cdot \mathbf{T}^{\text{init}} \cdot \mathbf{T}_i  \quad \text{if} \ i \in [s, N].
\vspace{-1mm}
\end{equation}
After each \ac{GICP} step at $i$, \(\mathbf{T}_i^{\text{ICP}}\) is also applied to subsequent point clouds to provide better initial guesses.
Since point clouds prior to index \( s \) may remain misaligned and there could be accumulated errors, we perform a \emph{backward} pass from \(i=N\) down to \(s\). This yields:
\begin{equation}
\mathbf{T}'_i = \prod_{k=i}^{N} \mathbf{T}_k^{\text{ICP,rev}} \cdot \prod_{j=s}^{i} \mathbf{T}_j^{\text{ICP}} \cdot \mathbf{T}^{\text{init}} \cdot \mathbf{T}_i.
\vspace{-1mm}
\end{equation}
Working in both directions---similar to ``zipping up'' two maps---improves overall alignment consistency and compensates the trajectory drift.

During scan-to-map registration, points with low $\epsilon_g$ (permanent) have higher weight in \ac{GICP}, while points with high $\epsilon_g$ (ephemeral) have less influence.  
\begin{equation}
    \mathbf{T}^{\text{ICP}} = \underset{\mathbf{T}}{\mathrm{argmin}} \sum\limits_{i} \left( 1 - \epsilon_{g,i} \right) \lVert \mathbf{d}_i(\mathbf{T}) \rVert^2_{\mathbf{\Sigma}_i},
\label{eq:4}
\end{equation}
\(\epsilon_{g,i}\) is the global ephemerality of the $i$th map point, and \(\mathbf{d}_i(\mathbf{T})\) is the residual error under transformation \(\mathbf{T}\).
Detailed explanation of \(\epsilon_{g}\) will be provided in \secref{subsec:dynamic_removal} and \secref{subsec:map_update}.
As the system incorporates more sessions, the $\epsilon_g$ distribution in the lifelong map becomes more reliable, steadily improving alignment robustness.

\subsection{Dynamic Object Removal}
\label{subsec:dynamic_removal}

\begin{figure}[!t]
    \centering
    \includegraphics[width=0.95\linewidth]{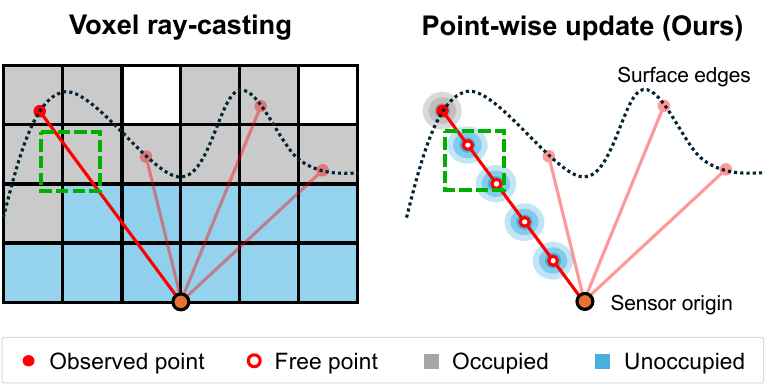}
    \caption{When voxelizing space, as indicated by the green square, the occupied area is inflated and errors can occur when a single voxel mixes static and dynamic points. Our method updates point-wise ephemerality based on ray information, enabling more precise removal of dynamic objects.}
    \label{fig:method_dynamic}
    \vspace{-3mm}
\end{figure}

The dynamic object removal module first aggregates all scans in the aligned session to form \(\mathcal{M}_{\mathcal{S}'_{t}} = \cup_{i=1}^{N} \mathbf{T}'_i \mathcal{P}_i\). Its goal is to produce a \emph{cleaned} session map \( \hat{\mathcal{M}}_{S'_t} \) by discarding dynamic points from \(\mathcal{M}_{\mathcal{S}'_{t}}\).

Treating each scan as a set of rays, we iteratively update the local ephemerality $\epsilon_{l}$ of points in \(\mathcal{M}_{\mathcal{S}'_{t}}\).  
A local ephemerality of a point should decrease if it is near an \emph{occupied} space (the endpoint of a ray) and increase if it lies in \emph{free} space (the path along a ray).
We propagate ephemerality across space by modeling a function of the distance $x$ from a ray:
\begin{equation}
\resizebox{0.91\linewidth}{!}{$
f(x)= \left\{
\begin{array}{lll}
\min \Bigl(\alpha \cdot \bigl(1 - \exp\{-x^2/{\sigma_\mathbf{o}}^2\}\bigr) + \beta,\ \alpha\Bigr) 
& \text{if} & \mathbf{o}_i \in \mathcal{O}_i \\[7pt]
\max \Bigl(\alpha \cdot \bigl(1 + \exp\{-x^2/{\sigma_\mathbf{f}}^2\}\bigr) - \beta,\ \alpha\Bigr) 
& \text{if} & \mathbf{f}_i \in \mathcal{F}_i \\
\end{array}
\right.
$}
\label{eq:5}
\end{equation}
\(\mathcal{O}_i\) is the set of endpoints in scan \(\mathcal{P}_i\) and \(\mathcal{F}_i\) consists of sampled points along the rays that approximate free space \cite{zhong2023shine}. 
\(\alpha\) and \(\beta\) are scale parameters (fixed to 0.5 and 0.1, respectively), while \(\sigma_{\mathbf{o}}\) and \(\sigma_{\mathbf{f}}\) are standard deviations.

Starting from an initial value \(\epsilon_{l} = 0.5\), each point in \(\mathcal{O}_i\) or \(\mathcal{F}_i\) retrieves its k-nearest neighbors in \(\mathcal{M}_{\mathcal{S}'_t}\) and updates their $\epsilon_{l}$ via Bayesian inference:
\begin{equation}
\epsilon_{\mathrm{l,new}} = \frac{f(x) \cdot \epsilon_{\mathrm{l,prev}}}{f(x) \cdot \epsilon_{\mathrm{l,prev}} + (1 - f(x)) \cdot (1 - \epsilon_{\mathrm{l,prev}})}.
\label{eq:6}
\end{equation}

As shown in \figref{fig:method_dynamic}, leveraging the actual rays rather than voxelization prevents the artificial inflation of occupied regions and leads to more precise ephemerality propagation.  
After several updates, points in \(\mathcal{M}_{\mathcal{S}'_t}\) with \(\epsilon_{l} < \tau_l\) are deemed static and pushed into \(\hat{\mathcal{M}}_{S'_t}\).

\subsection{Long-term Map Update}
\label{subsec:map_update}

\begin{figure}[!t]
    \centering
    \includegraphics[width=0.95\linewidth]{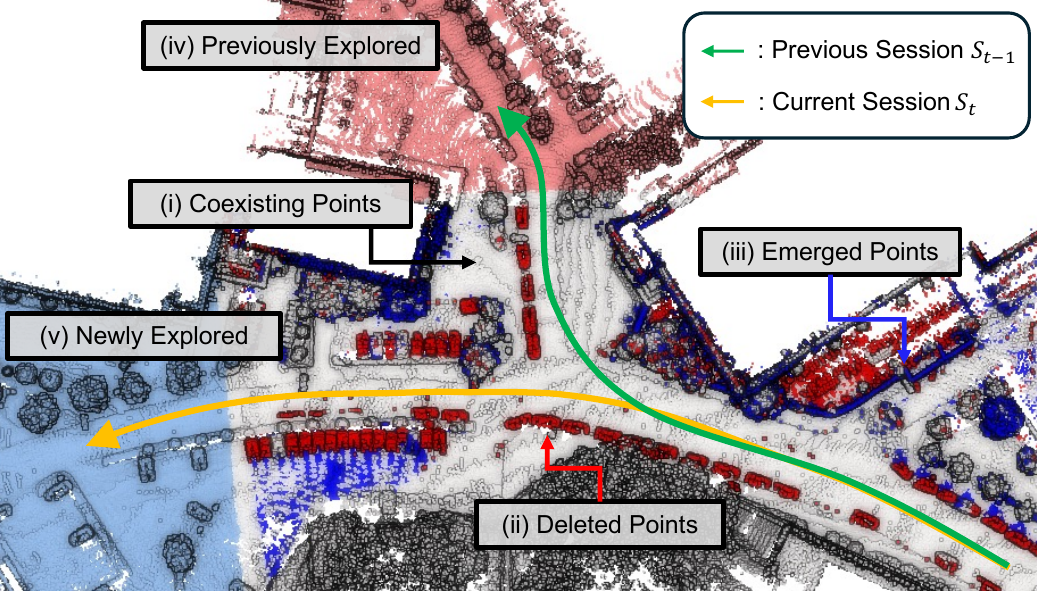}
    \caption{
    In our map update module, points are classified into five categories. 
    Coexisting points ($\mathcal{C}_t$) are shown in grey. 
    Deleted points ($\mathcal{D}_t$) are red if they truly disappeared and pink if they belong to previously visited regions only. 
    Emerged points ($\mathcal{E}_t$) are blue if newly added and sky blue if observed only in the current session. 
    Each category follows a specific update strategy for robust map maintenance.
    }
    \label{fig:method_c}
    \vspace{-3mm}
\end{figure}

The map update module merges the previous lifelong map \(\mathcal{M}_{t-1}\) with the cleaned session map \(\hat{\mathcal{M}}_{S'_t}\) to form the new lifelong map \(\mathcal{M}_{t}\):
\[
\mathcal{M}_{t} = \mathcal{M}_{t-1} \cup \hat{\mathcal{M}}_{S'_t}.
\]
It then classifies the points into five categories based on their nearest neighbors (NNs), as shown in \figref{fig:method_c}: 
(i) Coexisting points \( \mathcal{C}_t \),
(ii) Deleted points \( \mathcal{D}_t \) from \( \mathcal{M}_{t-1} \),
(iii) Emerged points \( \mathcal{E}_t \) in \( \hat{\mathcal{M}}_{S'_t} \).
Points in either \( \mathcal{D}_t \) or \( \mathcal{E}_t \) are further classified into:
(iv) Previously explored points, and 
(v) Newly explored points, if they are located in areas not explored by both the previous and current sessions.

For each point in $\mathcal{C}_t$, we update its $\epsilon_{g}$ via Bayesian inference using the previous $\epsilon_{g}$ (from $\mathcal{M}_{t-1}$) and the current $\epsilon_{l}$ (from $\hat{\mathcal{M}}_{S'_t}$):
\begin{equation}
\epsilon_{\mathrm{g_t}} = 
\frac{\epsilon_{\mathrm{g_{t-1}}} \cdot \epsilon_{l_t}}
{\epsilon_{\mathrm{g_{t-1}}} \cdot \epsilon_{l_t} + (1 - \epsilon_{\mathrm{g_{t-1}}}) \cdot (1 - \epsilon_{l_t})}.
\label{eq:7}
\end{equation}

The set \(\mathcal{D}_t\) can include truly removed objects as well as spurious points from sensor noise or pose errors.  
To emphasize meaningful changes and suppress noise, we introduce an \emph{objectness} factor $\gamma$, which prioritizes object-like clusters.  
We find that the local density $\rho$ of points effectively distinguishes object-level changes from noise and leverage it to define $\gamma$.
\begin{equation}
\gamma_i = \rho_i^{1/3}, \quad \rho_i \propto \{ \mathbf{p}_j \in \mathcal{D} \mid \|\mathbf{p}_j - \mathbf{p}_i\| \leq r, j \neq i \} 
\label{eq:8}
\end{equation}


Each deleted point’s $\epsilon_{g}$ are updated through Bayesian inference between previous $\epsilon_{g}$ and $\gamma$, similar to \eqref{eq:7}:

\begin{equation}
\epsilon_{\mathrm{g_t}} = 
\frac{\epsilon_{\mathrm{g_{t-1}}} \cdot \gamma}
{\epsilon_{\mathrm{g_{t-1}}} \cdot \gamma + (1 - \epsilon_{\mathrm{g_{t-1}}}) \cdot (1 - \gamma)}.
\label{eq:9}
\end{equation}

Meanwhile, emerged points \(\mathcal{E}_t\) comprise both newly built structures and ephemeral objects, introducing uncertainty of their permanence.
We thus scale their local ephemerality by an uncertainty factor \(k\) along with the objectness factor:
\begin{equation}
    \epsilon_{\mathrm{g,i}} = k \cdot \bigl(2-\gamma\bigr) \cdot  \epsilon_{\mathrm{l,i}} \quad \forall i \in \mathcal{E}_t.
    \label{eq:10}
\end{equation}
Points in category (iv) inherit their previous $\epsilon_g$ unchanged, while those in category (v) initialize $\epsilon_g$ directly from their $\epsilon_l$.

Over multiple sessions, truly static points accumulate enough observations for their $\epsilon_g$ to decrease and stabilize, distinguishing them from highly ephemeral objects.  
Changes in $\mathcal{M}_{t}$ are recorded in a \emph{delta map} for session $\mathcal{S}_t$, which continuously tracks and logs changes (\(\Delta \epsilon_g\)).  
Unlike LT-mapper’s \emph{diff map} \cite{kim2022lt}, this delta map captures the magnitude of change, enabling fine-grained analysis of environmental variation.  
Finally, a \emph{static map} can be retrieved by filtering $\mathcal{M}_t$ with a user-defined threshold \(\tau_g\) on $\epsilon_g$.

\section{EXPERIMENT}
\label{sec:experiment}

\subsection{Experimental Setup}
We conduct both quantitative and qualitative evaluations for each module of our system.

\textbf{Multi-session Map Alignment.}
We use six sequences from \texttt{LT-ParkingLot} \cite{kim2022lt} and MulRan \cite{kim2020mulran} (\texttt{DCC01-03} and \texttt{KAIST01-03}), as each provides multiple overlapping routes recorded at sufficiently spaced time intervals. Each first session is used to build the base map; the remaining sessions serve as new data for alignment. We employ SC-LIO-SAM \cite{shan2020lio, kim2021scan} for mapping each session.  
Following previous works \cite{hu2024paloc, yang2024lifelong}, we evaluate alignment using Accuracy (AC), RMSE, and Chamfer Distance (CD). 
In detail, we establish point correspondences between two point clouds using the nearest neighbor search and get the inlier set using a distance threshold $\sigma_{\text{inlier}}$ (set to 0.5$\m$).
AC measures the ratio of inlier pairs, RMSE is their root mean squared distance, and CD is the bidirectional sum of average inlier distance.
As baselines, we compare against ICP-based map-to-map registration \cite{girardeau2016cloudcompare} and multi-session PGO in LT-mapper \cite{kim2022lt}.

\textbf{Dynamic Object Removal.} 
Following \cite{lim2021erasor, jia2024beautymap, duberg2024dufomap}, we evaluate three different sequences from the SemanticKITTI \cite{behley2019semantickitti} dataset, adopting the protocol in \cite{zhang2023dynamic}. We use Preservation Rate (PR) and Removal Rate (RR) \cite{lim2021erasor} at the point level without downsampling the ground truth map to ensure accuracy. We also report the F1 score, the harmonic mean of PR and RR.  
Baselines include state-of-the-art methods such as Removert \cite{kim2020remove}, ERASOR \cite{lim2021erasor}, DUFOMap \cite{duberg2024dufomap}, and BeautyMap \cite{jia2024beautymap}.

\textbf{Map Update.}
Since no labeled dataset exists for inter-session changes, we conduct a qualitative evaluation using the MulRan \cite{kim2020mulran} (\texttt{KAIST01, 02}) and HeLiPR \cite{jung2023helipr} (\texttt{KAIST04, 05}) datasets. With a four-year gap between these datasets, we can observe both short-term and long-term changes.

\subsection{Multi-session Map Alignment}

\begin{table}[!t]
\centering
\caption{Map alignment evaluation results}
\begin{adjustbox}{width=1.0\linewidth}
{
\begin{tabular}{clccc}
\toprule
\multirow{2}{*}{Sequence}       & \multicolumn{1}{c}{\multirow{2}{*}{Method}} & \multicolumn{3}{c}{Metrics} \\  \cmidrule{3-5} 
                                & \multicolumn{1}{c}{}                        & AC $\uparrow$   & RMSE $\downarrow$     & CD $\downarrow$      \\
\midrule
\multirow{3}{*}{\texttt{LT-ParkingLot}} 
                                & ICP \cite{girardeau2016cloudcompare}         & 0.962    & 0.117    & 0.194 \\  
                                & LT-mapper \cite{kim2022lt}                   & 0.968    & 0.121    & 0.175 \\
                                & ELite (Ours)                                 & \textbf{0.969}    & \textbf{0.090}    & \textbf{0.133} \\
\midrule
\multirow{3}{*}{\texttt{DCC}}            
                                & ICP \cite{girardeau2016cloudcompare}         & 0.692     & 0.204    & 0.272 \\
                                & LT-mapper \cite{kim2022lt}                   & 0.738     & 0.182    & 0.306 \\
                                & ELite (Ours)                                 & \textbf{0.942}     & \textbf{0.111}    & \textbf{0.162} \\
\midrule
\multirow{3}{*}{\texttt{KAIST}}          
                                & ICP \cite{girardeau2016cloudcompare}         & 0.641     & 0.218    & 0.256 \\
                                & LT-mapper \cite{kim2022lt}                   & 0.909     & 0.169    & 0.315 \\
                                & ELite (Ours)                                 & \textbf{0.963}     & \textbf{0.120}    & \textbf{0.184} \\
\bottomrule
\end{tabular}
}
\end{adjustbox}
\tiny{ }

\footnotesize
\raggedright
\quad Best performance in \textbf{bold}.
\label{tab:map_eval}
\vspace{-4mm}
\end{table}


\begin{figure}[!t]
    \centering
    \subfigure[LT-mapper \cite{kim2022lt}]{        
        \includegraphics[width=0.45\linewidth]{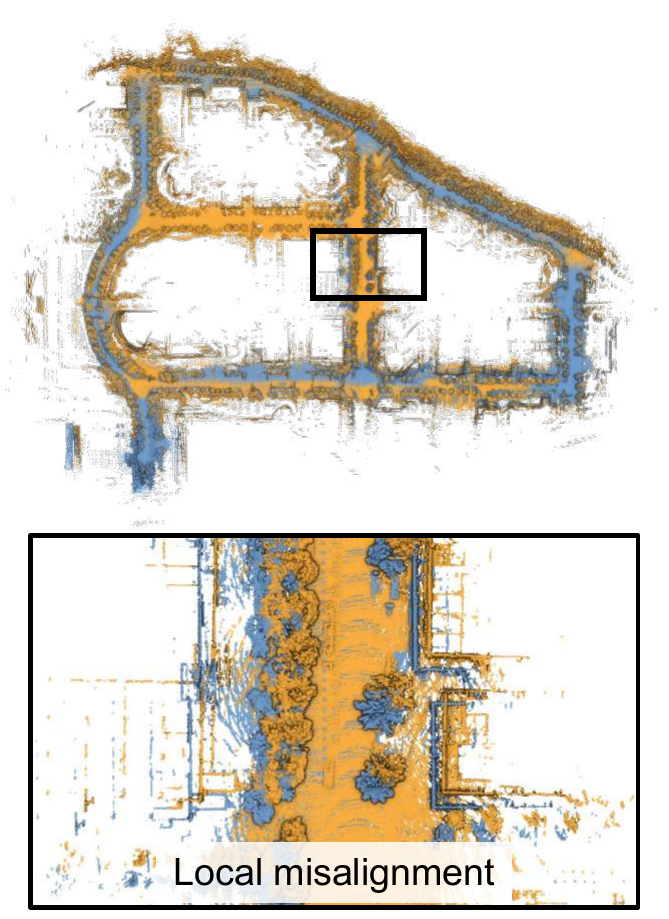}
        \label{fig:map_align_ltmapper}
    }
    \subfigure[Ours]{        
        \includegraphics[width=0.45\linewidth]{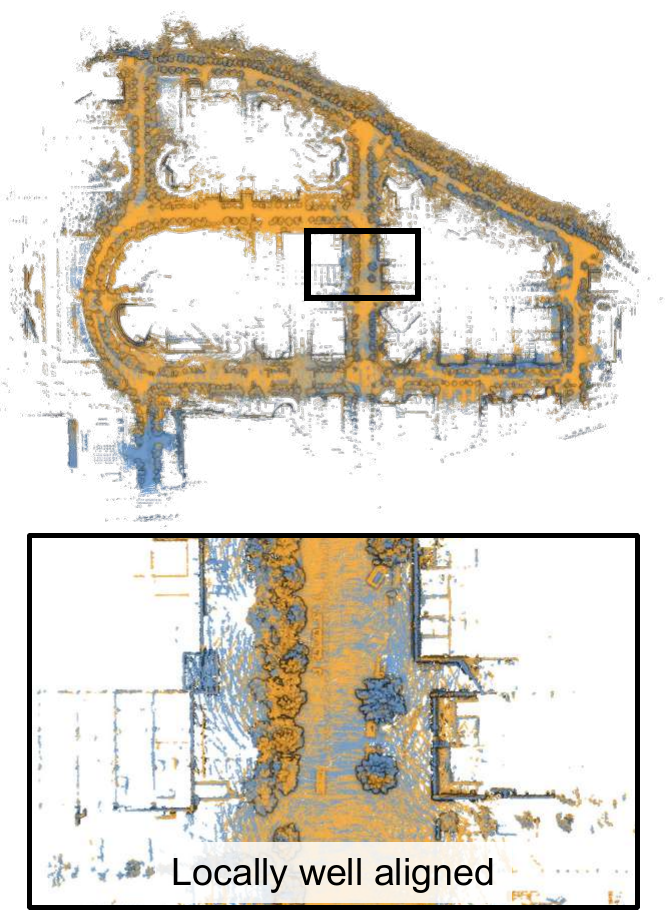}
        \label{fig:map_align_ours}
    }
    \caption{Qualitative comparison of multi-session map alignment on the \texttt{DCC} sequence in MulRan \cite{kim2020mulran}. Both methods exhibit globally consistent alignment, but our method demonstrates superior local consistency.}
    \label{fig:map_align}
    \vspace{-2mm}
\end{figure}

\begin{figure}[!t]
    \centering
    \subfigure[GICP \cite{koide2021voxelized}]{        
        \includegraphics[width=0.45\linewidth]{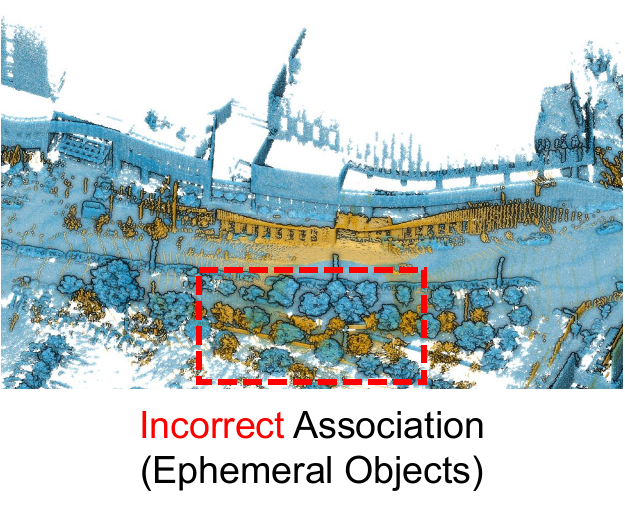}
        \label{fig:gicp}
    }
    \subfigure[Ours]{        
        \includegraphics[width=0.45\linewidth]{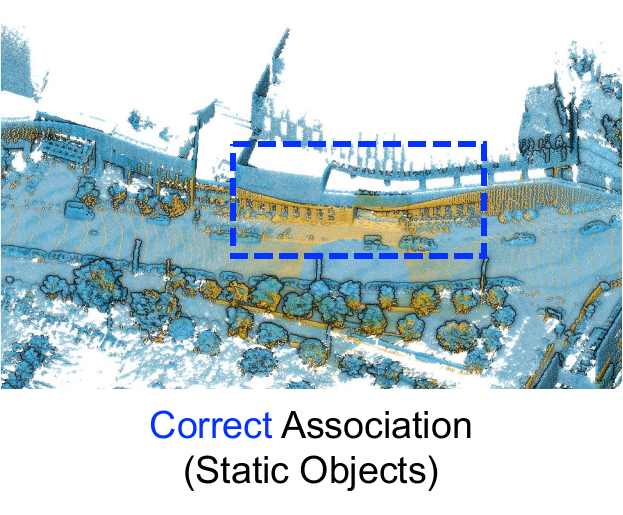}
        \label{fig:weighted_gicp}
    }
    \caption{Qualitative comparison on the \texttt{KAIST} sequence, showing the advantage of using $\epsilon_g$ as a weight for scan-to-map registration. By prioritizing long-term static structures, our method robustly aligns sessions even with significant time gaps or dynamic objects.}
    \label{fig:icp_comparison}
    \vspace{-5mm}
\end{figure}

\tabref{tab:map_eval} shows our map alignment performance, where our method consistently outperforms the baselines. While baselines can perform reasonably well in small-scale settings such as \texttt{LT-ParkingLot}, it struggles to register large-scale environments such as \texttt{DCC} and \texttt{KAIST}.  
\figref{fig:map_align} compares two session maps in the \texttt{DCC} sequence (\texttt{01} and \texttt{02}). LT-mapper \cite{kim2022lt}, which leverages multi-session PGO, achieves globally consistent but locally misaligned results. In contrast, our bidirectional registration yields both global and local consistency.  
\figref{fig:icp_comparison} shows the effectiveness of weighting scan-to-map registration with $\epsilon_g$. By assigning higher weights to static structures, our alignment module accurately registers sessions separated by huge time intervals or containing many dynamic objects.

\subsection{Dynamic Object Removal}

\begin{table}[t!]
\centering
\caption{Dynamic object removal results on SemanticKITTI}
\begin{adjustbox}{width=1.0\linewidth}
{
\begin{tabular}{c|l|rrc|rrc}
\toprule[0.8pt]
\multicolumn{1}{l|}{\multirow{2}{*}{Seq}} & \multicolumn{1}{c|}{\multirow{2}{*}{Method}}  & \multicolumn{3}{c|}{SuMa \cite{behley2019semantickitti,behley2018efficient}} & \multicolumn{3}{c}{KITTI Poses \cite{geiger2013vision}} \\ \cmidrule{3-8}
                                             &                                               & \multicolumn{1}{c}{PR $\uparrow$} & \multicolumn{1}{c}{RR $\uparrow$} & F1 $\uparrow$ & \multicolumn{1}{c}{PR $\uparrow$} & \multicolumn{1}{c}{RR $\uparrow$} & F1 $\uparrow$ \\
\midrule[0.6pt]
\multirow{6}{*}{\texttt{00}} & Removert \cite{kim2020remove}   & 99.55 & 41.14 & 58.22  & 99.24 & 41.42 & 58.44           \\
                    & ERASOR \cite{lim2021erasor}    & 70.23 & 98.49 & 81.98   & 65.99 & 98.32 & 78.98                    \\
                    & DUFOMap \cite{duberg2024dufomap}    & 98.63 & 98.66 & \textbf{98.64}  & 92.59 & 98.47 & 95.44           \\
                    & BeautyMap \cite{jia2024beautymap}   & 97.13 & 97.79 & 97.46 & 97.07 & 97.84 & \textbf{97.45}           \\
                    & ELite - 0.2 (Ours) & 97.30 & 98.74 & 98.02 & 93.22 & 98.55 & 95.81        \\
                    & ELite - 0.5 (Ours) & 98.54 & 98.28 & \underline{98.41} & 96.61 & 97.93 & \underline{97.27}        \\ 
                    \midrule
\multirow{6}{*}{\texttt{01}} & Removert \cite{kim2020remove}   & 98.27 & 39.47 & 56.32  & 98.43 & 39.85 & 56.73                    \\
                    & ERASOR \cite{lim2021erasor}    & 98.35 & 90.96 & 94.51    & 83.06 & 92.43 & 87.49                    \\
                    & DUFOMap \cite{duberg2024dufomap}    & 98.94 & 93.93 & \underline{96.37}  & 98.97 & 93.52 & 96.17        \\
                    & BeautyMap \cite{jia2024beautymap}    & 99.30 & 92.37 & 95.71 & 99.20 & 90.20 & 94.48                    \\
                    & ELite - 0.2 (Ours) & 94.70 & 97.84 & 96.24 & 95.18 & 97.86 & \underline{96.50}        \\
                    & ELite - 0.5 (Ours) & 96.72 & 96.52 & \textbf{96.62} & 97.15 & 96.62 & \textbf{96.89}  \\ 
                    \midrule
\multirow{6}{*}{\texttt{02}} & Removert \cite{kim2020remove}   & 96.69 & 35.26 & 51.68  & 97.18 & 35.34 & 51.83                    \\
                    & ERASOR \cite{lim2021erasor}      & 50.79 & 92.26 & 65.51  & 51.63 & 93.83 & 66.61                    \\
                    & DUFOMap \cite{duberg2024dufomap}    & 68.61 & 89.29 & 77.60  & 70.78 & 89.49 & 79.04                    \\
                    & BeautyMap \cite{jia2024beautymap}   & 83.43 & 84.66 & 84.04 & 82.07 & 90.61 & \underline{86.13}        \\
                    & ELite - 0.2 (Ours) & 80.71 & 91.21 & \underline{85.64} & 80.44 & 90.84 & 85.32        \\
                    & ELite - 0.5 (Ours) & 84.03 & 89.70 & \textbf{86.77} & 83.62 & 89.21 & \textbf{86.32}  \\ 
\bottomrule[0.8pt]
\end{tabular}
}
\end{adjustbox}
\tiny{ }

\footnotesize
\raggedright
Best performance in \textbf{bold}, second best \underline{underlined}.
\label{tab:semantic_kitti}
\vspace{-5mm}
\end{table}

\begin{figure}[!t]
    \centering
    \subfigure[Raw Map]{        
        \includegraphics[width=0.45\linewidth]{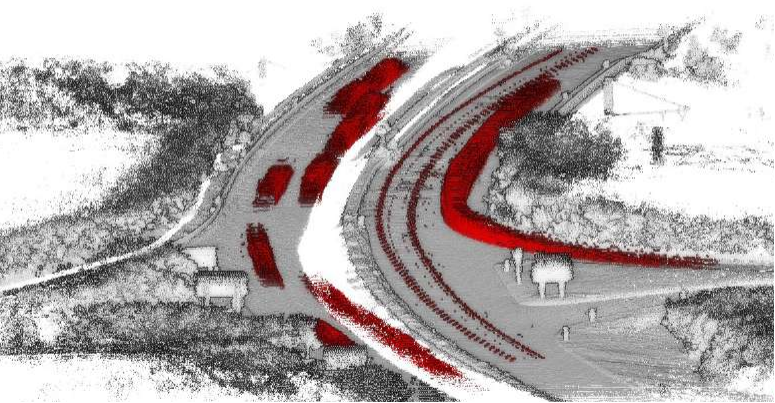}
        \label{fig:result_gt}
    }
    \subfigure[Removert \cite{kim2020remove}]{        
        \includegraphics[width=0.45\linewidth]{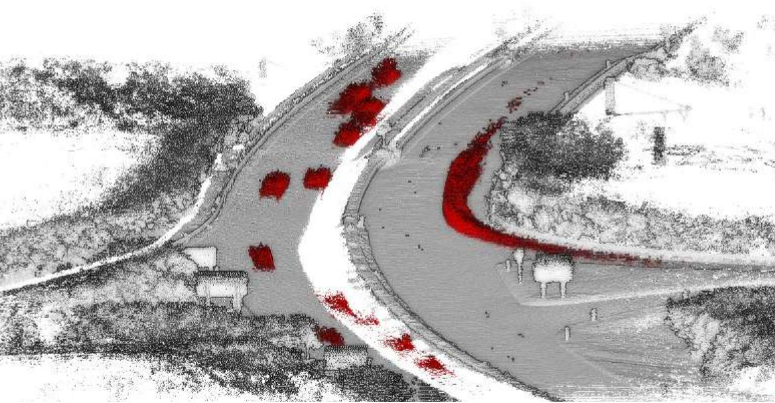}
        \label{fig:result_removert}
    } \\
    \subfigure[ERASOR \cite{lim2021erasor}]{        
        \includegraphics[width=0.45\linewidth]{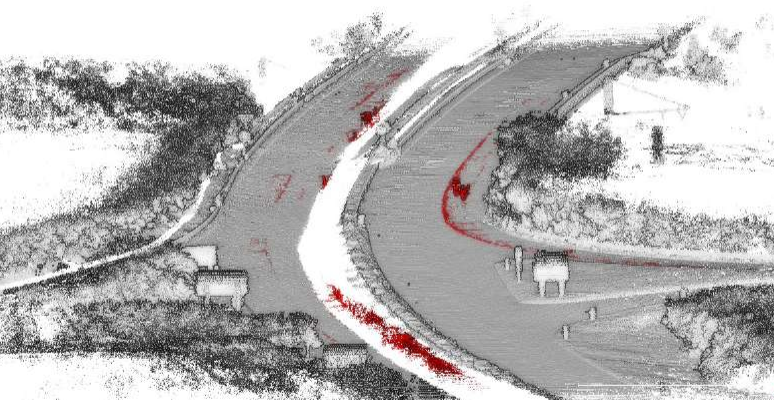}
        \label{fig:erasor}
    }
    \subfigure[DUFOMap \cite{duberg2024dufomap}]{        
        \includegraphics[width=0.45\linewidth]{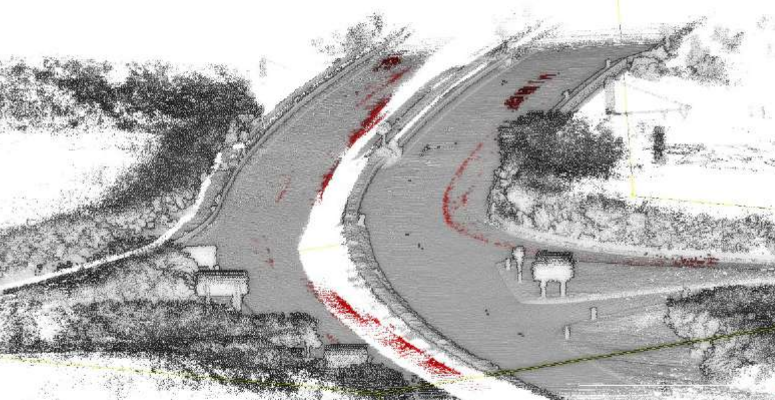}
        \label{fig:result_dufomap}
    } \\
    \subfigure[BeautyMap \cite{jia2024beautymap}]{        
        \includegraphics[width=0.45\linewidth]{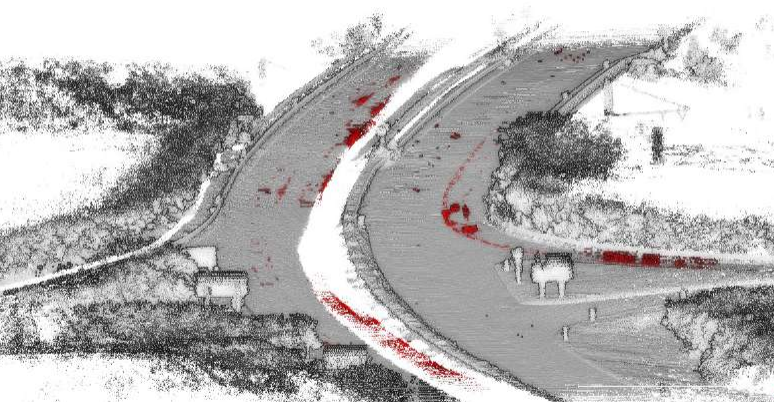}
        \label{fig:result_beautymap}
    }
    \subfigure[Ours]{        
        \includegraphics[width=0.45\linewidth]{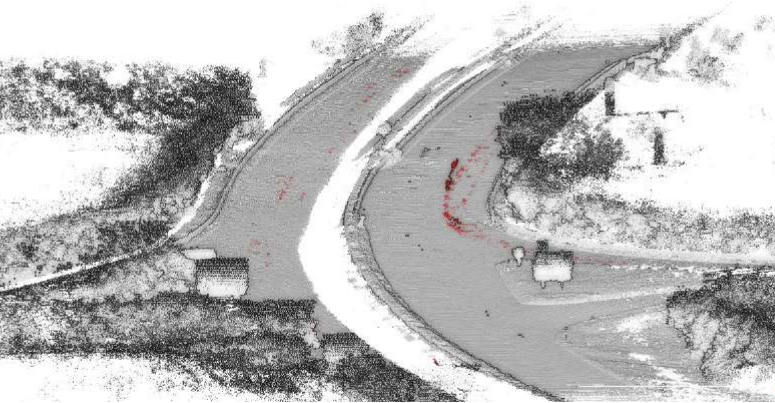}
        \label{fig:result_ours}
    }
    \caption{Qualitative comparison of dynamic object removal in the SemanticKITTI \cite{behley2019semantickitti} \texttt{01} sequence. The ground truth dynamic points in (a) are shown in red. Figures (b)-(f) depict the cleaned maps produced by each method, with red points indicating remaining dynamic objects.}
    \label{fig:result_dynamic_removal}
    \vspace{-5mm}
\end{figure}

\tabref{tab:semantic_kitti} and \figref{fig:result_dynamic_removal} illustrate the performance of various dynamic object removal methods. We report results for two thresholds of $\epsilon_l$: 0.2 and 0.5. Our approach at $\tau_{l} = 0.5$ achieves the highest or comparable performance on most sequences. Lowering the threshold to $\tau_{l} = 0.2$ significantly increases RR with only a moderate sacrifice in PR, showing the flexibility to prioritize either aggressive removal of dynamic objects or better preservation of static points.

We also evaluate two different pose estimation sources: KITTI poses \cite{geiger2013vision} and SuMa poses \cite{behley2018efficient} from SemanticKITTI \cite{behley2019semantickitti}. As noted in \tabref{tab:semantic_kitti}, all methods perform worse when using KITTI poses, indicating the importance of accurate pose estimation. Spatial quantization-based methods degrade further with less accurate poses, while our point-wise iterative update remains comparatively robust.

\subsection{Map Update}

\begin{figure*}[!t]
    \centering
    \includegraphics[width=0.95\linewidth]{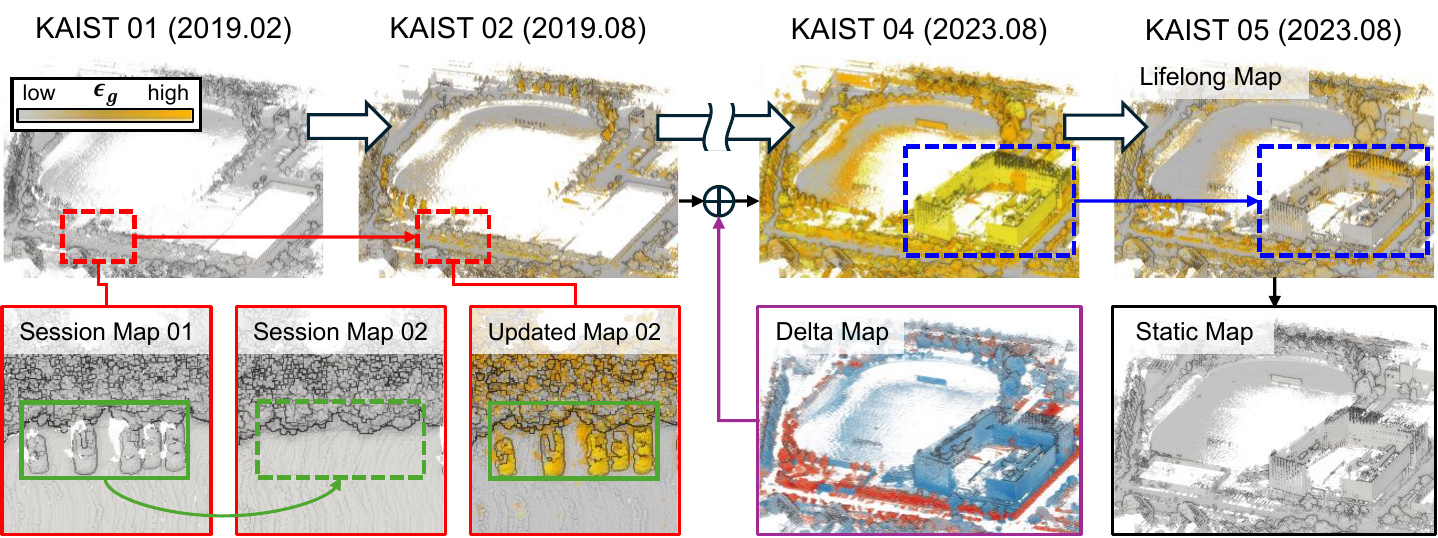}
    \caption{Sample scene from the \texttt{KAIST} sequences of MulRan \cite{kim2020mulran} and HeLiPR \cite{jung2023helipr}. The red box highlights short-term changes, such as parked cars whose ephemerality increases over time. The blue box highlights long-term changes, including a newly constructed building between \texttt{02} and \texttt{04}. While its initial ephemerality is slightly high, indicating potential transience, it decreases as more updates confirm its permanence.}
    \label{fig:eph_change}
    \vspace{-5mm}
\end{figure*}

\figref{fig:eph_change} shows the results of our map update module. The top row depicts the evolving lifelong map over several sessions. After multiple updates, transient objects like parked cars (red box) exhibit increasing ephemerality, aligning with their dynamic nature. Meanwhile, newly built structures (blue box) see a gradual decrease in ephemerality, reflecting long-term permanence. Thus, two-stage ephemerality propagation effectively distinguishes static from ephemeral objects.  
ELite can also produce a static map by filtering points whose $\epsilon_g$ is below a user-defined threshold $\tau_g$. A smaller $\tau_g$ yields a map retaining only long-term static structures, whereas a higher $\tau_g$ preserves moderately ephemeral objects (e.g., currently parked cars). This threshold acts as a convenient way to tailor the static map for specific needs.

\begin{figure}[!t]
    \centering
    \subfigure[Diff map \cite{kim2022lt}]{        
        \includegraphics[width=0.45\linewidth]{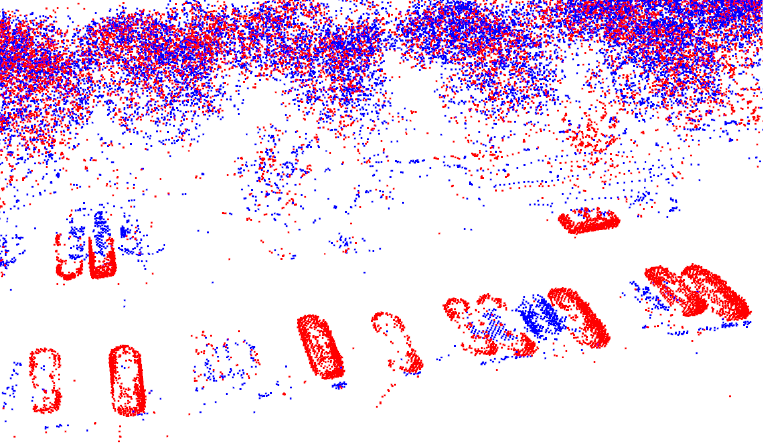}
        \label{fig:density_a}
    }
    \subfigure[Delta map (ours)]{        
        \includegraphics[width=0.45\linewidth]{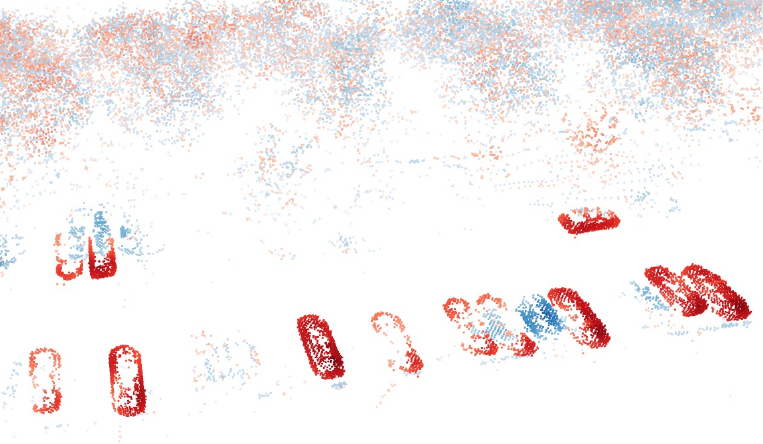}
        \label{fig:density_b}
    }
    \caption{
    Qualitative comparison of inter-session change representations. The diff map \cite{kim2022lt} uses a binary appearance/disappearance scheme, treating true changes and measurement noise equally. In contrast, our delta map includes an objectness factor $\gamma$ (darker colors correspond to higher values), which assigns greater weights to object-level changes and suppresses noise.
    }
    \label{fig:density}
    \vspace{-3mm}
\end{figure}

\figref{fig:density} compares the diff map \cite{kim2022lt} to our delta map.
While the diff map provides only binary information about point appearance or disappearance, our delta map introduces an objectness factor $\gamma$ to accurately separate meaningful changes from artifacts, thus enabling more robust map updates.

\subsection{Potential Downstream Tasks}
\begin{figure}[!t]
    \centering
    \includegraphics[width=0.95\linewidth]{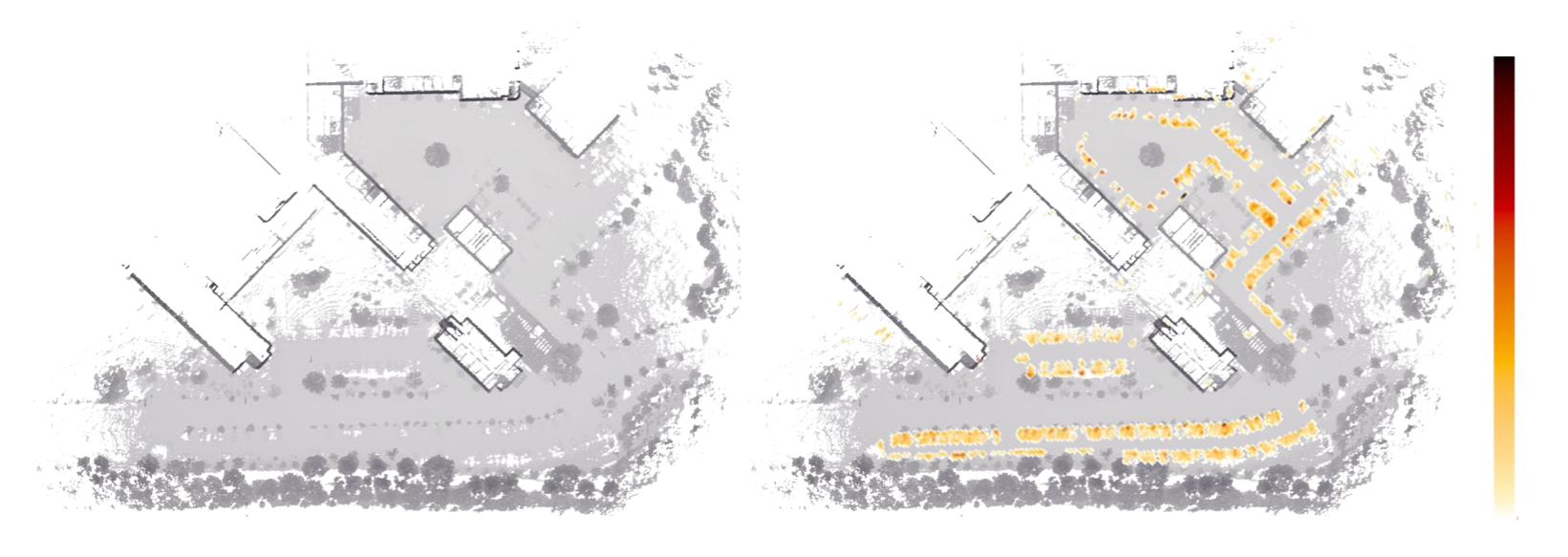}
    \caption{
    \textsl{Left}: Static map constructed from the \texttt{LT-ParkingLot} dataset. 
    \textsl{Right}: Heatmap of frequently changing points after updates from six sessions, overlaid on the static map. Higher heatmap values indicate a stronger likelihood of ephemeral objects, aiding in navigation or planning.
    }
    \label{fig:applications}
    \vspace{-3mm}
\end{figure}

Using delta maps, we can create a heatmap indicating how frequently each region undergoes change. As depicted in \figref{fig:applications}, ephemeral objects tend to yield high heatmap values, identifying areas prone to frequent changes---a potentially valuable insight for optimizing robot navigation.  
Moreover, with enough updates, time-domain analysis similar to \cite{krajnik2017fremen} can be applied, and this can be integrated into \eqref{eq:7} to reflect more region-specific dynamics. We plan to explore this direction in our future work.

\section{CONCLUSION}
\label{sec:conclusion}

We present ELite, a LiDAR-based lifelong mapping framework that distinguishes map elements beyond static and dynamic using two-stage ephemerality. 
ELite integrates map alignment, dynamic object removal, and updates into a cohesive system, enhancing performance by propagating ephemerality across modules. 
By providing an extended representation for map changes, ELite enables meaningful updates and supports applications such as static map construction and spatial analysis. 
Real-world validation confirms its accuracy and reliability, making ELite a valuable tool for long-term robot operation.






\newpage

\balance
\small
\bibliographystyle{IEEEtranN} 
\bibliography{string-short,references}

\end{document}